%% file: emnlp2019.tex
\documentclass[11pt,a4paper]{article}

\usepackage[hyperref]{emnlp-ijcnlp-2019}
\usepackage{times}
\usepackage{latexsym}
\usepackage{url}
\usepackage{float}
\usepackage{amsmath}
\usepackage{amssymb}
\usepackage{microtype}
\usepackage{subfigure}
\usepackage{booktabs}% for professional tables
\usepackage{amsthm}
\usepackage{amsfonts}
\usepackage{arydshln}
\usepackage{multirow}
\usepackage{multicol}
\usepackage{comment}

\usepackage{tcolorbox}

\aclfinalcopy % Uncomment this line for the final submission

\def\KL{\textsf{KL}} %KL function symbol
\input{subtex/mlVecMat.tex}
\usepackage{algorithm}
\usepackage{algorithmic}
\renewcommand{\algorithmicrequire}{\textbf{Input:}}
\renewcommand{\algorithmicensure}{\textbf{Output:}}

\begin{document}

\title{Implicit Deep Latent Variable Models for Text Generation}
\author{Le Fang$^\dagger$,
    ~~Chunyuan Li$^\mathsection$,
    ~~Jianfeng Gao$^\mathsection$, 
    ~~Wen Dong$^\dagger$,
    ~~Changyou Chen$^\dagger$\\
    $^\dagger$University at Buffalo,~~
    $^\mathsection$Microsoft Research, Redmond \\
    \texttt{ \{lefang, wendong, changyou\}@buffalo.edu }\\ 
    \texttt{ \{chunyl, jfgao\}@microsoft.com }
 }
\date{}
\maketitle

\begin{abstract}
Deep latent variable models (LVM) such as variational auto-encoder (VAE) have recently played an important role in text generation. 
One key factor is the exploitation of smooth latent structures to guide the generation. However, the representation power of VAEs is limited due to two reasons: (1) the Gaussian assumption is often made on the variational posteriors; and meanwhile (2) a notorious ``posterior collapse'' issue occurs.
% impacts and significance of latent code can be severely weakened , due to a restricted assumption on encoding distributions and
In this paper, we advocate sample-based representations of variational distributions for natural language, leading to implicit latent features, which can provide flexible representation power compared with Gaussian-based posteriors. 
We further develop an LVM to directly match the aggregated posterior to the prior. 
It can be viewed as a natural extension of VAEs with a regularization of maximizing mutual information, mitigating the ``posterior collapse'' issue.
% We analyze the inherent deficiency of original VAE objectives, which performs the same prior regularization for all data inputs. 
% We correct this by regularizing the aggregated latent space. This simple modification essentially incorporates mutual information into the optimization and remedies the ``posterior collapse'' issue.
% The proposed implicit auto-encoder stands as a natural extension of current mainstream VAEs with significant performance enhancement. 
We demonstrate the effectiveness and versatility of our models in various text generation scenarios, including language modeling, unaligned style transfer, and dialog response generation.
The source code to reproduce our experimental results is available on GitHub\footnote{\url{https://github.com/fangleai/Implicit-LVM}}.
\end{abstract}

\section{Introduction}
Deep latent variable models (LVM) such as variational auto-encoder (VAE) \cite{kingma2013auto, rezende2014stochastic} are successfully applied for many natural language processing tasks, including 
% especially in text generation under various scenarios and specifications. 
%Successful applications include 
language modeling \cite{bowman2015generating, miao2016neural}, dialogue response generation \cite{zhao2017learning}, controllable text generation \cite{hu2017toward} and neural machine translation \cite{shah_generative} etc. One advantage of VAEs is the flexible distribution-based latent representation. It captures holistic properties of input, such as style, topic, and high-level linguistic/semantic features, which further guide the generation of diverse and relevant sentences.

% \begin{figure}
% \centering
% \subfigure[VAE]{
% \label{fig:VAE}
% \includegraphics[width=2.9cm]{figs/vae.png}}
% \subfigure[Implicit VAE]{
% \label{fig:implicit}
% \includegraphics[width=3.1cm]{figs/implicit.png}}
% \caption{Latent spaces from VAE and implicit VAE.}
% \label{fig:comparison}
% \end{figure}

However, the representation capacity of VAEs is restrictive due to two reasons.
The first reason is rooted in the assumption of variational posteriors, which usually follow spherical Gaussian distributions with diagonal co-variance matrices. It has been shown that an approximation gap generally exists between the true posterior and the best possible variational posterior in a restricted family~\cite{cremer2018inference}. Consequently, the gap may militate against learning an optimal generative model, as its parameters may be always updated based on sub-optimal posteriors~\cite{kim2018semi}. The second reason is the so-called {\em posterior collapse} issue, which occurs when learning VAEs with an auto-regressive decoder~\cite{bowman2015generating}. It produces undesirable outcomes: the encoder yields meaningless posteriors that are very close to the prior, while the decoder tends to ignore the latent codes in generation~\cite{bowman2015generating}. 
Several attempts have been made to alleviate this issue~\cite{bowman2015generating,higgins2017beta,zhao2017infovae,fu2019cyclical,he2019lagging}.

These two seemingly unrelated issues are studied independently. 
In this paper, we argue that the posterior collapse issue is partially due to the restrictive Gaussian assumption, as it limits the optimization space of the encoder/decoder in a given distribution family. 
$(\RN{1})$
To break the assumption, we propose to use sample-based representations for natural language, thus leading to implicit latent features. Such a representation is much more expressive than Gaussian-based posteriors.
$(\RN{2})$
This implicit representation allows us to extend VAE and develop new LVM that further mitigate the posterior collapse issue. 
% We observe the inherent deficiency of original VAE objectives that the KL divergence regularization term matches each posterior distribution independently to the same prior distribution and presents no correlation between latent codes of different data. 
% An unfortunate but very likely case is that all KL regularization terms vanish toward zero, causing posterior collapse at same time for all inputs. 
It represents all the sentences in the dataset as posterior samples in the latent space, and matches the aggregated posterior samples to the prior distribution. Consequently, latent features are encouraged to cooperate and behave diversely to capture meaningful information for each sentence. 

However, learning with implicit representations faces one challenge: it is intractable to evaluate the KL divergence term in the objectives. We overcome the issue by introducing a conjugate-dual form of the KL divergence \cite{rockafellar1966extension, NIPS2018_8177}. It facilitates learning via training an auxiliary dual function.
%
% In this paper, we present effective implicit deep latent variable models which utilize more flexible sample-based representation to capture essential language information for downstream text generation. 
%
The effectiveness of our models is validated by producing consistently supreme results on a broad range of generation tasks, including language modeling, unsupervised style transfer, and dialog response generation. 
% We note our models as better alternatives of current VAEs in text generation tasks as far as we have tested. 
%

%is provided in Supplementary Materials along with submission, and will be made publicly available on GitHub\footnote{\url{https://github.com}}.

\section{Preliminaries}
% We review preliminary concepts about implicit distribution and VAEs for text generation.

% \subsection{Implicit distributions}
% %
% Implicit models have been successfully applied to generative modelling \cite{GAN, makhzani2015adversarial} and variational inference \cite{huszar2017variational, Yin2018SemiImplicitVI}. Generally saying, implicit distributions are probability models whose density function may be intractable, but there is a way to sample from them and calculate gradients with respect to model parameters to update the distributions. A popular example of implicit models are stochastic generative networks where samples from a simple base distribution, such as Gaussian, are transformed by a deep neural network. Such distributions can flexibly represent a wide range of probability distributions, including degenerate distributions which may not even have a continuous density \cite{huszar2017variational}.

% \subsection{VAE for text generation}
When applied to text generation, VAEs~\cite{bowman2015generating} consist of two parts, a generative network (decoder) and an inference network (encoder).
Given a training dataset $\mathcal{D}=\{\xv_{i}\}_{i=1}^{\left|\mathcal{D}\right|}$, where $\xv_{i} = [x_{1i}, \cdots, x_{Ti} ]$ represents $i$th sentence of length $T$.
Starting from a prior distribution $p(\zv)$, VAE generates a sentence $\xv$ using the deep generative network $p_{\thetav}(\xv|\zv)$, where $\thetav$ is the network parameter. Therefore, the joint distribution $p_{\thetav}(\xv,\zv)$ is defined as $p(\zv)p_{\thetav}(\xv|\zv)$.
The prior $p(\zv)$ is typically assumed as a standard multivariate Gaussian.
% $p_{\thetav}(\xv,\zv)=p_{\thetav}(\xv|\zv)p(\zv)$.
Due to the sequential nature of natural language, the decoder $p_{\thetav}(\xv|\zv)$ takes an auto-regressive form $p_{\thetav}(\xv|\zv)=\prod_{t=1}^{T}p_{\thetav}(x_{t}|x_{<t},\zv)$.
The goal of model training is to maximize the marginal data log-likelihood $\mathbb{E}_{\xv\sim \mathcal{D}}\text{log} p_{\thetav}(\xv)$.

%However, the integral in likelihood is generally intractable and variational inference \cite{jordan1999introduction} is considered. 

However, it is intractable to perform posterior inference. An $\phiv$-parameterized encoder is introduced to approximate $p_{\thetav}(\zv|\xv)\propto p_{\thetav}(\xv|\zv)p(\zv)$ with a variational distribution $q_{\phiv}(\zv|\xv)$. Variational inference is employed for VAE learning, yielding following evidence lower bound (ELBO):
\begin{align}
  &  \mathbb{E}_{\xv \sim \Dcal} \text{log} p_{\thetav}(\xv) \nonumber \geq \Lcal_1 = \Lcal_2 ~~\text{with}~~ \\
  \Lcal_1  & = \Lcal_E +  \Lcal_{R}~~\text{where}~~\label{eq:ELBO-1} \\
   & ~~\Lcal_E  = \mathbb{E}_{\xv\sim \Dcal}\left[\mathbb{E}_{\zv\sim q_{\phiv}(\zv|\xv)}\text{log}p_{\thetav}(\xv|\zv) \right] \\
 & ~~~\Lcal_{R} = \mathbb{E}_{\xv\sim \Dcal}\left[-\KL\left(q_{\phiv}(\zv|\xv)\parallel p(\zv)\right)\right] 
 \label{eq:kl_term}\\
 \Lcal_2  & = \mathbb{E}_{\xv\sim \Dcal}\text{log}p_{\thetav}(\xv) +  \Lcal_{G}~~\text{where}~~ \\
 & 
 ~~\Lcal_{G} =
 -\mathbb{E}_{\xv\sim \Dcal}\KL\left(q_{\phiv}(\zv|\xv)\parallel p_{\thetav}(\zv|\xv)\right)\label{eq:ELBO-2}
\end{align}
Note that $\Lcal_1$ and $\Lcal_2$ provide two different views for the VAE objective:
\begin{itemize}
    \item $\Lcal_1$ consists of a reconstruction error term $\Lcal_E$ and a KL divergence regularization term $\Lcal_R$. With a strong auto-regressive decoder $p_{\thetav}(\xv|\zv)$, the objective tends to degenerate all encoding distribution $q_{\phiv}(\zv|\xv)$ to the prior, causing $\Lcal_R \rightarrow 0$, \ie the posterior collapse issue.
    \vspace{-2mm}
    \item 
    $\Lcal_2$ indicates that VAE requests a flexible encoding distribution family to minimize the approximation gap $\Lcal_G$ between the true posterior and the best possible encoding distribution. This motivates us to perform more flexible posterior inference with implicit representations.
\end{itemize}

\section{The Proposed Models}\label{sec:method}
We introduce a sample-based latent representation for natural language, and develop two models that leverage its advantages.  (1) When replacing the Gaussian variational distributions with  sample-based distributions in VAEs, we derive {\em implicit VAE} (iVAE). (2) We further extend VAE to maximize mutual information between latent representations and observed sentences, leading to a variant termed as iVAE$_{\text{MI}}$.

\subsection{Implicit VAE}
\begin{figure*}[t!]
\centering
\includegraphics[width=15.4cm]{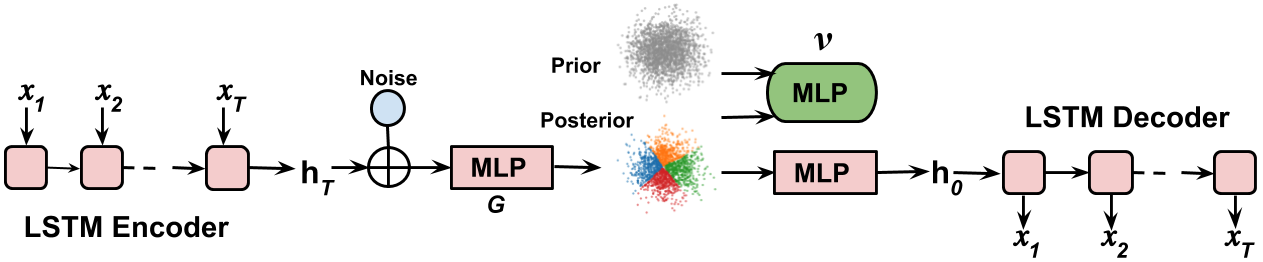}
\vspace{-0mm}
\caption{Illustration of the proposed implicit LVMs. $\nu(\xv, \zv)$ is used in iVAE, and $\nu(\zv)$ is used in iVAE$_{\text{MI}}$. In this example, the piror is $p(\zv)=\Ncal(0,1)$, the sample-based aggregated posterior $q(\zv) = \int q_{\phiv}(\zv| \xv) q(\xv)d \xv$ for four observations is shown, where the posterior $q_{\phiv}(\zv| \xv)$ for each observation is visualized in a different color.}
\label{fig:lm_model}
\vspace{-3mm}
\end{figure*}

% An implicit VAE (iVAE) learns sample-based representation. Specifically, 
%
\paragraph{Implicit Representations}
Instead of assuming an explicit density form such as Gaussian, we define a sampling mechanism to represent $q_{\phiv}(\zv|\xv)$ as a set of sample $\{ \zv_{\xv,i}\}_{i=1}^M$, through the encoder as
\begin{align}
\zv_{\xv,i} = G_{\phiv}(\xv, \epsilon_i), \epsilon_i \sim q(\epsilon)
\label{eq:implicit}
\end{align}
where the $i$-th sample is drawn from a neural network $G_{\phiv}$ that takes $(\xv, \epsilon_i)$ as input; $q(\epsilon)$ is a simple distribution such as standard Gaussian. It is difficult to naively combine the random noise $\epsilon$ with the sentence $\xv$ (a sequence of discrete tokens) as the input of $G_{\phiv}$. Our solution is to concatenate noise $\epsilon_i$ with hidden representations $\hv$ of $\xv$. $\hv$ is generated using a LSTM encoder, as illustrated in Figure~\ref{fig:lm_model}.
% or its intermediate equivalent in $G$ to compute a code sample.

% \paragraph{Connection to vanilla autoencoder}
% Vanilla autoencoder learns a point estimate of latent code to reconstruct a given data point \cite{dai2015semi}. It grants full flexibility on latent codes, but the latent space is not fully utilized due to point-mass rather than distributional representations. Vanilla autoencoder therefore lacks regularization and is prone to overfitting.

\paragraph{Dual form of KL-divergence} 
Though theoretically promising, the implicit representations in~\eqref{eq:implicit} render difficulties in optimizing the KL term $\Lcal_{R}$ in \eqref{eq:kl_term}, as the functional form is no longer tractable with implicit $q_{\phiv}(\zv|\xv)$. We resort to evaluating its dual form based on Fenchel duality theorem \cite{rockafellar1966extension, NIPS2018_8177}:
\begin{align}
 &\KL\left(q_{\phiv}(\zv|\xv) \parallel p(\zv)\right) \label{eq:KL-dual} \\
=&\text{max}_{\nu} \mathbb{E}_{\zv\sim q_{\phiv}(\zv|\xv)}\nu_{\psiv}(\xv,\zv)-\mathbb{E}_{\zv\sim p(\zv)} \text{exp}(\nu_{\psiv}(\xv,\zv)),\nonumber
\end{align}
where $\nu_{\psiv}(\xv,\zv)$ is an auxiliary dual function, parameterized by a neural network with weights $\psiv$. By replacing the KL term with this dual form, the implicit VAE has the following objective:
\begin{align}
 \mathcal{L}_{\text{iVAE}} \nonumber
& = \mathbb{E}_{\xv \sim \Dcal} \mathbb{E}_{\zv\sim q_{\phiv}(\zv|\xv)}\text{log} p_{\thetav}(\xv|\zv) \\
&  -\mathbb{E}_{\xv \sim \Dcal} \mathbb{E}_{\zv\sim q_{\phiv}(\zv|\xv)}\nu_{\psiv}(\xv,\zv) \nonumber \\
&  +\mathbb{E}_{\xv \sim \Dcal} \mathbb{E}_{\zv\sim p(\zv)} \text{exp}(\nu_{\psiv}(\xv,\zv)),  \label{eq:I-VAE}
\end{align}

\paragraph{Training scheme} 
Implicit VAE inherits the end-to-end training scheme of VAEs with extra work on training the auxiliary network  $\nu_{\psiv}(\xv,\zv)$: %\cy{It might help to write out the training process for VAE as well. ----$>$ maybe, but I think VAE process is clearly known for all, but depend on you, I can add it afterwards}:
\begin{itemize}
\item Sample a mini-batch of $\xv_i\sim\mathcal{D}$, $\epsilon_i\sim q(\epsilon)$, and generate $\zv_{\xv_i,\epsilon_i} = G(\xv_i, \epsilon_i; \phiv)$; Sample a mini-batch of $\zv_i\sim p(\zv)$.
\item Update ${\psiv}$ in $\nu_{\psiv}(\xv,\zv)$ to maximize 
\begin{align}
\hspace{-3mm}
\sum_{i}\nu_{\psiv}(\xv_i,\zv_{\xv_i,\epsilon_i})-\sum_{i}\text{exp}(\nu_{\psiv}(\xv_i,\zv_i))
\label{eq:ivae_dual}
\end{align}
\item Update parameters $\{\phiv, \thetav\}$ to maximize
\begin{align}
\hspace{-3mm}
\sum_{i}\text{log}p_{\thetav}(\xv_i|\zv_{\xv_i,\epsilon_i})-\sum_{i}\nu_{\psiv}(\xv_i,\zv_{\xv_i,\epsilon_i})
\label{eq:ivae_prime}
\end{align}
\end{itemize}
In practice, we implement $\nu_{\psiv}(\xv,\zv)$ with a multilayer perceptron (MLP), which takes the concatenation of $\hv$ and $\zv$. In another word, the auxiliary network distinguishes between $(\xv, \zv_{\xv})$ and  $(\xv, \zv)$, where $\zv_{\xv}$ is drawn from the posterior and $\zv$  is drawn from the prior, respectively. We found the  MLP-parameterized auxiliary network converges faster than LSTM encoder and decoder~\cite{hochreiter1997long}. This means that the auxiliary network practically provides an accurate approximation to the KL regularization $\Lcal_{R}$.

\subsection{Mutual Information Regularized iVAE}
% We further develop a variant that strongly mitigates the posterior collapse issue. 
%
% The implicit distribution relax 

It is noted the inherent deficiency of the original VAE objective in \eqref{eq:kl_term}: the KL divergence regularization term matches each posterior distribution independently to the same prior.
% and presents no correlation between latent codes of different sentences. 
%
% This may not be an issue in other domains like images
This is prone to posterior collapse in text generation, due to a strong auto-regressive decoder $p_{\thetav}(\xv|\zv)$. When sequentially generating $x_t$, the model learns to solely rely on the ground-truth $[x_1, \cdots, x_{t-1}]$, and ignore the dependency from $\zv$~\cite{fu2019cyclical}.
It results in the learned variational posteriors $q_{\phiv}(\zv|\xv)$ to exactly match $p(\zv)$, without discriminating data $\xv$. 

To better regularize the latent space, we propose to replace $\Lcal_{R} = \mathbb{E}_{\xv\sim \Dcal}\left[-\KL\left(q_{\phiv}(\zv|\xv)\parallel p(\zv)\right)\right]$ in~\eqref{eq:kl_term}, with the following KL divergence:
\begin{equation}
-\KL\left(q_{\phiv}(\zv)\parallel p(\zv)\right), \label{eq:agg_KL_MI}
\end{equation}
where  $q_{\phiv}(\zv) = \int q(\xv)q_{\phiv}(\zv|\xv)d\xv$ is the {\em aggregated posterior}, $q(\xv)$ is the empirical data distribution for the training dataset $\mathcal{D}$. The integral is estimated by ancestral sampling in practice, {\it i.e.} we first sample $\xv$ from dataset and then sample $\zv\sim q_{\phiv}(\zv|\xv)$. 

In~\eqref{eq:agg_KL_MI}, variational posterior is regularized as a whole $q_{\phiv}(\zv)$, encouraging posterior samples from different sentences to cooperate to satisfy the objective.
% and behave diversely to capture meaningful information for each data. 
It implies a solution that each sentence is represented as a local region in the latent space, the aggregated representation of all sentences match the prior; This avoids the degenerated solution from ~\eqref{eq:kl_term} that the feature representation of individual sentence spans over the whole space.

\paragraph{Connection to mutual information}
The proposed latent variable model coincides with \cite{zhao2018unsupervised, zhao2017infovae} where mutual information is introduced into the optimization, based on the following decomposition result (Please see detailed proof in Appendix \ref{decomp}):
%
%\[-\KL\left(q_{\phiv}(\zv)\parallel p(\zv)\right)=-\mathbb{E}_{\xv}\KL\left(q_{\phiv}(\zv|\xv)\parallel p(\zv)\right)+I(\xv,\zv),\]
%
\begin{align}
-\KL\left(q_{\phiv}(\zv)\parallel p(\zv)\right) & = I(\xv,\zv) \nonumber \\
&-\mathbb{E}_{\xv}\KL\left(q_{\phiv}(\zv|\xv)\parallel p(\zv)\right),\nonumber
\end{align}
where $I(\xv,\zv)$ is the mutual information between $\zv$ and $\xv$ under the joint distribution $q_{\phiv}(\xv,\zv)=q(\xv)q_{\phiv}(\zv|\xv)$. Therefore, the objective in \eqref{eq:agg_KL_MI} also maximizes the mutual information between individual sentences and their latent features. We term the new LVM objective as  iVAE$_{\text{MI}}$: 
\begin{align}
 \Lcal_{ \text{iVAE}_{\text{MI}} } 
 & = \mathbb{E}_{\xv \sim \Dcal} \mathbb{E}_{\zv\sim q_{\phiv}(\zv|\xv)} \text{log} p_{\thetav}(\xv|\zv) \nonumber
 \\
 &~~~~ - \KL\left( q_{\phiv}(\zv)\parallel p(\zv) \right) \label{eq:KL_MI}
\end{align}

\paragraph{Training scheme}
Note that the aggregated posterior $q_{\phiv}(\zv)$ is also a sample-based distribution. 
Similarly, we evaluate \eqref{eq:KL_MI} through its dual form:
\begin{align}
 &\KL\left(q_{\phiv}(\zv) \parallel p(\zv)\right) \label{eq:KL-dual-mi} \\
=&\text{max}_{\nu} \mathbb{E}_{\zv\sim q_{\phiv}(\zv)}\nu_{\psiv}(\zv)-\mathbb{E}_{\zv\sim p(\zv)} \text{exp}(\nu_{\psiv}(\zv)).\nonumber
\end{align}
Therefore, iVAE$_{\text{MI}}$ in \eqref{eq:KL_MI} can be written as:
\begin{align}
 \mathcal{L}_{\text{iVAE}_{\text{MI}}}  \label{eq:I-VAE-MI}
& = \mathbb{E}_{\xv \sim D} \mathbb{E}_{\zv\sim q_{\phiv}(\zv|\xv)}\text{log}p_{\thetav}(\xv|\zv) \\
&  -\mathbb{E}_{\zv\sim q_{\phiv}(\zv)}\nu_{\psiv}(\zv)+\mathbb{E}_{\zv\sim p(\zv)} \text{exp}(\nu_{\psiv}(\zv)), \nonumber
\end{align}
where the auxiliary network $\nu_{\psiv}(\zv)$ is parameterized as a neural network. Different from iVAE, $\nu_{\psiv}(\zv)$ in iVAE$_{\text{MI}}$ only takes posterior samples as input. 
The training algorithm is similar to iVAE in Section 3.1, except a different auxiliary network $\nu_{\psiv}(\zv)$. In Appendix \ref{I-VAE-MI}, we show the full algorithm of iVAE$_{\text{MI}}$.

We illustrate the proposed methods in Figure~\ref{fig:lm_model}. Note that both iVAE and iVAE$_{\text{MI}}$ share the same model architecture, except a different auxiliary network $\nu$.

\section{Related Work}
\vspace{-1mm}
\subsection{Solutions to posterior collapse}
Several attempts have been made to alleviate the posterior collapse issue. The KL annealing scheme in language VAEs has been first used in ~\cite{bowman2015generating}. An effective cyclical KL annealing schedule is used in~\cite{fu2019cyclical}, where the KL annealing process is repeated multiple times. KL term weighting scheme is also adopted in $\beta$-VAE~\cite{higgins2017beta} for disentanglement. On model architecture side, dilated CNN was considered to replace auto-regressive LSTMs for decoding~\cite{yang2017improved}. The bag-of-word auxiliary loss was proposed to improve the dependence on latent representations in generation~\cite{zhao2017learning}. More recently, lagging inference proposes to aggressively update encoder multiple times before a single decoder update~\cite{he2019lagging}.  Semi-amortized VAE refines variational parameters from an amortized encoder per instance with stochastic variational inference~\cite{kim2018semi}. 

All these efforts utilize the Gaussian-based forms for posterior inference. Our paper is among the first ones to attribute posterior collapse issue to the restrictive Gaussian assumption, and advocate more flexible sample-based representations.

\subsection{Implicit Feature Learning}
Sample-based distributions, as well as implicit features, have been widely used in representation learning~\cite{donahue2017adversarial,li2017alice}. 
Vanilla autoencoders learn point masses of latent features rather than their distributions.
Adversarial variational Bayes introduces an auxiliary discriminator network like GANs~\cite{GAN, makhzani2015adversarial} to learn almost arbitrarily distributed latent variables \cite{mescheder2017adversarial,pu2017adversarial}. We explore the similar spirit in the natural language processing (NLP) domain. Amortized MCMC and particle based methods are introduced for LVM learning in~\cite{li2017approximate,pu2017vae,chen2018continuous}. 
Coupled variational Bayes~\cite{NIPS2018_8177} emphasizes an optimization embedding, {\em i.e.}, a flow of particles, in a general setting of non-parametric variational inference. It also utilizes similar dual form with auxiliary function $\nu_{\psiv}(\xv,\zv)$ to evaluate KL divergence.  Adversarially regularized autoencoders \cite{makhzani2015adversarial, kim2017adversarially} use similar objectives with iVAEs, in the form of a reconstruction error plus a specific regularization evaluated with implicit samples. Mutual information has also been considered into regularization in \cite{zhao2018unsupervised, zhao2017infovae} to obtain more informative representations. 

Most previous works focus on image domain. It is largely unexplored in NLP. Further, the auto-regressive decoder renders an additional challenge when applying implicit latent representations. Adversarial training with samples can be empirically unstable, and slow even applying recent stabilization techniques in GANs \cite{Arjovsky2017WassersteinG, gulrajani2017wasserstein}. To the best of our knowledge, this paper presents the first to effectively apply implicit feature representations, to NLP.

\section{Experiments}\label{sec:exp}
% In this section, the effectiveness of our methods are validated by producing consistently state-of-the-art metrics on a broad range of text generation tasks, including language modeling, unaligned style transfer and dialog response generation. % Case studies cover text generation under various scenarios, demonstrating that our models perform better in producing meaningful and diverse representations, with posterior collapse issue significantly alleviated as well.
In this section, the effectiveness of our methods is validated by largely producing supreme metrics on a broad range of text generation tasks under various scenarios. We note that when presenting our results in terms of perplexity and lower bounds, our evaluations are approximated. Thus they are not directly comparable to other methods. For a fair comparison, more advanced evaluation methods such as the annealing importance sampling should be adapted, which we leave as future work. Other evaluation metrics, {\it e.g.}, the BLEU score, are fair to compare.

\vspace{-0mm}
\subsection{Language Modeling}\label{subsec:lm}
% Language modeling is a fundamental task to mimic the language generation process with RNNs conditioned on latent representations.

\paragraph{Datasets.} 
We consider three public datasets, the Penn Treebank ($\mathtt{PTB}$) \cite{marcus1993building, bowman2015generating}, $\mathtt{Yahoo}$, and  $\mathtt{Yelp}$  corpora~\cite{yang2017improved, he2019lagging}.
$\mathtt{PTB}$ is a relatively small dataset with  sentences of varying lengths, whereas $\mathtt{Yahoo}$ and $\mathtt{Yelp}$ contain larger amounts of long sentences. Detailed statistics of these datasets are shown in Table~\ref{table:lm_datasets}. 

\begin{table}[t!]
\begin{centering}
\begin{adjustbox}{scale=0.9,tabular=c|c|c|c,center}
\toprule
Dataset & $\mathtt{PTB}$ & $\mathtt{Yahoo}$ & $\mathtt{Yelp}$ \tabularnewline
\midrule
num. sents. & 42,000 & 100,000 & 100,000 \tabularnewline
min. len. & 2 & 21 & 21 \tabularnewline
max. len. & 78 & 201 & 201 \tabularnewline
avg. len. & 21.9 & 79.9 & 96.7 \tabularnewline
\bottomrule
\end{adjustbox}
\end{centering}
\caption{Statistics of datasets for language modeling.}
\label{table:lm_datasets}
\vspace{2mm}
\end{table}

\paragraph{Settings} We implement both encoder and decoder as one-layer LSTM. As illustrated in Figure~\ref{fig:lm_model}, concatenation of the last hidden state of the encoder LSTM and a Gaussian noise is fed into a MLP to draw a posterior sample. When decoding, a latent sample is fed into a MLP to produce the initial hidden state of the decoder LSTM. A KL-term annealing scheme is also used \cite{bowman2015generating}. We list more detailed hyper-parameters and architectures in Appendix \ref{train_lm}.

\paragraph{Baselines.} We compare with several state-of-the-art VAE language modeling methods, including (1) VAEs with a monotonically KL-annealing schedule~\cite{bowman2015generating}; (2) $\beta$-VAE \cite{higgins2017beta}, VAEs with a small penalty on KL term scaled by $\beta$; (3) SA-VAE \cite{kim2018semi}, mixing instance-specific variational inference with amortized inference; (4) Cyclical VAE \cite{fu2019cyclical} that periodically anneals the KL term; (5) Lagging VAE, which aggressively updates encoder multiple times before a single decoder update. 
% Note that some methods are essentially parallel to implicit VAE with techniques that can be directly merged with ours, while the benchmark shows how they work independently.

% \paragraph{Performance}

\begin{table}[t]
\begin{centering}
\begin{adjustbox}{scale=0.9,tabular=c|c|c|c|c|c,center}
\toprule
Methods & -ELBO\! $\downarrow$ & PPL\! $\downarrow$ & KL\! $\uparrow$  & MI\! $\uparrow$ & AU\! $\uparrow$ \tabularnewline
\hline
\multicolumn{6}{c}{Dataset: $\mathtt{PTB}$}\tabularnewline
\hline
VAE & 102.6 & 108.26 & 1.08 & 0.8 & 2 \tabularnewline
$\beta$(0.5)-VAE & 104.5 & 117.92 & \textbf{7.50} & 3.1 & 5 \tabularnewline
SA-VAE & 102.6 & 107.71 & 1.23 & 0.7 & 2 \tabularnewline
Cyc-VAE & 103.1 & 110.50 & 3.48 & 1.8 & 5 \tabularnewline
iVAE & \textbf{87.6} & \textbf{54.46} & 6.32 & \textbf{3.5} & \textbf{32}\tabularnewline
iVAE$_{\text{MI}}$ & \textbf{87.2} & \textbf{53.44} & \textbf{12.51} & \textbf{12.2} & \textbf{32}\tabularnewline
\hline
\multicolumn{6}{c}{Dataset:  $\mathtt{Yahoo}$}\tabularnewline %, \cite{he2019lagging} for models except ours
\hline 
VAE & 328.6 & 61.21 & 0.0 & 0.0 & 0\tabularnewline
$\beta$(0.4)-VAE & 328.7 & 61.29 & 6.3 & 2.8 & 8 \tabularnewline
SA-VAE & 327.2 & 60.15 & 5.2 & 2.7 & 10 \tabularnewline
Lag-VAE & 326.7 & 59.77 & 5.7 & 2.9 & 15 \tabularnewline
iVAE & \textbf{309.5} & \textbf{48.22} & \textbf{8.0} & \textbf{4.4} & \textbf{32}\tabularnewline
iVAE$_{\text{MI}}$ & \textbf{309.1} & \textbf{47.93} & \textbf{11.4} & \textbf{10.7} & \textbf{32}\tabularnewline
\hline
\multicolumn{6}{c}{Dataset: $\mathtt{Yelp}$}\tabularnewline %, \cite{he2019lagging} for models except ours
\hline 
VAE & 357.9 & 40.56 & 0.0 & 0.0 & 0 \tabularnewline
$\beta$(0.4)-VAE & 358.2 & 40.69 & 4.2 & 2.0 & 4 \tabularnewline
SA-VAE & 355.9 & 39.73 & 2.8 & 1.7 & 8 \tabularnewline
Lag-VAE & 355.9 & 39.73 & 3.8 & 2.4 & 11 \tabularnewline
iVAE & \textbf{348.2} & \textbf{36.70} & \textbf{7.6} & \textbf{4.6} & \textbf{32}\tabularnewline
iVAE$_{\text{MI}}$ & \textbf{348.7} & \textbf{36.88} & \textbf{11.6} & \textbf{11.0} & \textbf{32}\tabularnewline
\bottomrule
\end{adjustbox}
\end{centering}
\caption{Language modeling on three datasets. }
\label{table:languagemodeling}
\vspace{-2mm}
\end{table}

\paragraph{Evaluation metrics.} Two categories of metrics are used to study VAEs for language modeling: 
\begin{itemize}
    \item To characterize the modeling ability of the observed sentences, we use the negative ELBO as the sum of reconstruction loss and KL term, as well as perplexity (PPL).
    \vspace{-2mm}
    \item Compared with traditional neural language models, VAEs has its unique advantages in feature learning. To measure the quality of learned features, we consider 
    (1) KL: $\KL\left(q_{\phiv}(\zv|\xv)\parallel p(\zv)\right)$;
    (2) Mutual information (MI) $I(\xv,\zv)$ under the joint distribution $q_{\phiv}(\xv,\zv)$; 
    (3) Number of active units (AU) of latent representation. The activity of a latent dimension $z$ is measured as $A_{\zv}$$=\text{Cov}_{\xv}\left(\mathbb{E}_{\zv\sim q(\zv|\xv)}\left[z\right]\right)$, which is defined as active if $A_{\zv}$$>0.01$.  
\end{itemize}

The evaluation of implicit LVMs is unexplored in language models, as there is no analytical forms for the KL term.  We consider to evaluate both $\KL\left(q_{\phiv}(\zv)\parallel p(\zv)\right)$ and $\KL\left(q_{\phiv}(\zv|\xv)\parallel p(\zv)\right)$ by training a fully connected $\nu$ network following Eq. \eqref{eq:KL-dual} and \eqref{eq:KL-dual-mi}. 
To avoid the inconsistency between $\nu(\xv,\zv)$ and $\nu(\zv)$ networks due to training, we train them using the same data and optimizer in every iteration. We evaluate each distribution $q_{\phiv}(\zv|\xv)$ with 128 latent samples per $\xv$. We report the KL terms until their learning curves are empirically converged. Note that our evaluation is different from exact PPL in two ways: (1) It is a bound of $\log p(\xv)$ (similar to all text VAE model in the literature); (2) the KL term in the bound is estimated using samples, rather than a closed form. 
%Note that we recognize our metrics as the limiting approximation values we can approach towards the precise ones, due to the intractable nature of implicit representations. Therefore we don't claim state-of-the-art performance when compared with other methods evaluated in common settings.

We report the results in Table \ref{table:languagemodeling}.
A better language model would pursue a lower negative ELBO (also lower reconstruction errors, lower PPL), and make sufficient use of the latent space ({\it i.e.}, maintain relatively high KL term, higher mutual information and more active units). 
Under all these metrics, the proposed iVAEs achieve much better performance. The posterior collapse issue is largely alleviated as indicated by the improved KL and MI values, especially with iVAE$_{\text{MI}}$ which directly takes mutual information into account.

\begin{table}[t]
\begin{centering}
\begin{adjustbox}{scale=0.8,tabular=c|cc|cc|cc,center}
\toprule
\multirow{2}{*}{Dataset} & \multicolumn{2}{c|}{$\mathtt{PTB}$} & \multicolumn{2}{c|}{$\mathtt{Yahoo}$} & \multicolumn{2}{c}{$\mathtt{Yelp}$}\tabularnewline
\cline{2-7} 
 & Re.\! $\downarrow$ & Abs.\! $\downarrow$ & Re.\! $\downarrow$ & Abs.\! $\downarrow$ & Re.\! $\downarrow$ & Abs.\! $\downarrow$ \tabularnewline
\midrule
VAE/$\beta$- & 1.0 & 1.3 & 1.0 & 5.3 & 1.0 & 5.7 \tabularnewline
SA-VAE & 5.5 & 7.1 & 9.9 & 52.9 & 10.3 & 59.3 \tabularnewline 
Lag-VAE & - & - & 2.2 & 11.7 & 3.7 & 21.4 \tabularnewline
iVAEs & 1.4 & 1.8 & 1.3 & 6.9 & 1.3 & 7.5 \tabularnewline
\bottomrule
\end{adjustbox}
\end{centering}
\caption{Total training time in hours: absolute time and relative time versus VAE.}
\label{table:lm_traintime}
\end{table}

The comparison on training time is shown in Table \ref{table:lm_traintime}. iVAE and iVAE$_{\text{MI}}$ requires updating an auxiliary network, it spends $30\%$ more time than traditional VAEs. This is more efficient than SA-VAE and Lag-VAE.

% \footnote{The ``Lag-VAE'' cell of $\mathtt{PTB}$ is actually ``Cyc-VAE''.}.
% owing to the end-to-end learning where extra price is only paid to train a fully connected neural network.

\begin{table}[t] \footnotesize{}
	%\centering
    \begin{tcolorbox} 
      \hspace{-5mm}
    \begin{tabular}{l l}
%$\xv_{1}$ & \textbf{in new york the company declined comment} \tabularnewline
$t=$0 & in new york the company declined comment \tabularnewline
$t=$0.1 & in new york the company declined comment\tabularnewline
$t=$0.2 & in new york the transaction was suspended\tabularnewline
$t=$0.3 & in the securities company said yesterday\tabularnewline
$t=$0.4 & in other board the transaction had disclosed\tabularnewline
$t=$0.5 & other of those has been available\tabularnewline
$t=$0.6 & both of companies have been unchanged\tabularnewline
$t=$0.7 & both men have received a plan to restructure\tabularnewline
$t=$0.8 & and to reduce that it owns\tabularnewline
$t=$0.9 & and to continue to make prices\tabularnewline
$t=$1 & and they plan to buy more today\tabularnewline
%$\xv_{2}$ & \textbf{and they plan to buy more today} \tabularnewline  
	\end{tabular}
	\end{tcolorbox}
	\caption{Interpolating latent representation.}
	\label{table:interpolation}
	\vspace{2mm}
\end{table}

\paragraph{Latent space interpolation.} 
One favorable property of VAEs \cite{bowman2015generating, zhao2018unsupervised} is to provide smooth latent  representations that capture sentence semantics. We demonstrate this by interpolating two latent feature, each of which represents a unique sentence. Table \ref{table:interpolation} shows the generated examples. We take two sentences $\xv_{1}$ and $\xv_{2}$, and obtain their latent features as the sample-averaging results for $\zv_{1}$ and $\zv_{2}$, respectively, from the implicit encoder, and then greedily decode conditioned on the interpolated feature $\zv_{t}=\zv_{1}\cdot(1-t)+\zv_{2}\cdot t$ with $t$ increased from 0 to 1 by a step size of 0.1. It generates sentences with smooth semantic evolution.

\begin{table}[t!]
\begin{centering}
\begin{adjustbox}{scale=0.9,tabular=c|c|c,center}
\toprule
Model & Forward\! $\downarrow$ & Reverse\! $\downarrow$ \tabularnewline
\midrule 
VAE & 18494 & 10149 \tabularnewline
Cyc-VAE & 3390 & 5587 \tabularnewline
AE & 672 & 2589 \tabularnewline
$\beta$(0)-VAE & 625 & 1897 \tabularnewline
$\beta$(0.5)-VAE & 939 & 4078 \tabularnewline
SA-VAE & 341 & 10651 \tabularnewline
iVAE & \textbf{116} & \textbf{1520} \tabularnewline
iVAE$_{\text{MI}}$ & \textbf{134} & \textbf{1137} \tabularnewline
\bottomrule
\end{adjustbox}
\end{centering}
\caption{Forward and reverse PPL on $\mathtt{PTB}$.}
\label{table:forward_reverse}
\vspace{-0.0cm}
\end{table}

\paragraph{Improved Decoder.} 
One might wonder whether the improvements come from simply having a more flexible encoder during evaluation, rather than from utilizing high-quality latent features, and learning a better decoder. 
%
% Recall Figure \ref{fig:comparison}, the latent space learned from VAE appears to have holes and gaps, compared to the uniform dense space from an implicit VAE. 
%
We use $\mathtt{PTB}$ to confirm our findings. We draw samples from the prior $p(\zv)$, and greedily decode them using the trained decoder. The quality of the generated text is evaluated by an external library ``KenLM Language Model Toolkit'' \cite{Heafield-estimate} with two metrics \cite{kim2017adversarially}: (1) Forward PPL: the fluency of the generated text based on language models derived from the $\mathtt{PTB}$ training corpus; (2) Reverse PPL: the fluency of $\mathtt{PTB}$ corpus based on language model derived from the generated text, which measures the extent to which the generations are representative of the $\mathtt{PTB}$ underlying language model. For both the PPL numbers, the lower the better. We use n=5 for n-gram language models in ``KenLM''.

As shown in Table \ref{table:forward_reverse}, implicit LVMs outperform others in both PPLs, which confirms that the implicit representation can lead to better decoders. The vanilla VAE model performs the worst. This is expected, as the posterior collapse issue results in poor utilization of a latent space. % SA-VAE benefits from the stochastic refinements.
Besides, we can see that iVAE$_{\text{MI}}$ generates comparably fluent but more diverse text than pure iVAE, from the lower reverse PPL values. This is reasonable, due to the ability of iVAE$_{\text{MI}}$ to encourage diverse latent samples per sentence with the aggregated regularization in the latent space.

\subsection{Unaligned style transfer}
We next consider the task of unaligned style transfer, which represents a scenario to generate text with desired specifications. The goal is to control one style aspect of a sentence that is independent of its content. We consider non-parallel corpora of sentences, where the sentences in two corpora have the same content distribution but with different styles, and no paired sentence is provided. 

% We specifically consider transferring binary sentiment on $\mathtt{Yelp}$ restaurant reviews.

\paragraph{Model Extension.}
The success of this task depends on the exploitation of the distributional equivalence of content to learn sentiment independent content code and decoding it to a different style. 
To ensure such independence, we extend iVAE$_{\text{MI}}$ by adding a sentiment classifier loss to its objective (\ref{eq:I-VAE-MI}), similar to the previous style transfer methods~\cite{shen2017style, kim2017adversarially}.
Let $y$ be the style attribute, $\xv_{p}$ and $\xv_{n}$ (with corresponding features $\zv_{p}$, $\zv_{n}$) be sentences with positive and negative sentiments , respectively. The style classifier loss $\mathcal{L}_{\text{class}}(\zv_{p}, \zv_{n})$ is the cross entropy loss of a binary classifier. 

The classifier and encoder are trained adversarially: (1) the classifier is trained to distinguish latent features with different sentiments; (2) the encoder is trained to fool the classifier in order to remove distinctions of content features from sentiments. 
In practice, the classifier is implemented as a MLP. We implement two separate decoder LSTMs for clean sentiment decoding: one for positive $p(\xv|\zv,y=1)$, and one for negative sentiment $p(\xv|\zv,y=0)$. The prior $p(\zv)$ is also implemented as an implicit distribution, via transforming noise from a standard Gaussian through a MLP. Appendix \ref{train_st} lists more details. 
\paragraph{Baseline.}
We compare iVAE$_{\text{MI}}$ with a state-of-the-art unaligned sentiment transferer: the adversarially regularized autoencoder (ARAE) \cite{kim2017adversarially}, which directly regularizes the latent space with Wasserstein distance measure.
\begin{table} [t] \scriptsize
	%\centering
	
	\begin{tcolorbox} 
	\hspace{-5mm}
	\begin{tabular}{l l}
        {\bf Input}: & it was super dry and had a \textcolor{red}{weird}  \textcolor{red}{taste} to the entire \textcolor{red}{slice} . \\
        {\bf ARAE}: & it was \textcolor{blue}{super nice} and \textcolor{blue}{the owner} was \textcolor{blue}{super sweet and helpful} . \\
        {\bf iVAE$_{\text{MI}}$}: & it was \textcolor{red}{super tasty} and a good size with the best in the \textcolor{red}{burgh} . \\
         \hfill  \\
        
        {\bf Input}: & so i only had \textcolor{red}{half} of the regular \textcolor{red}{fries and my soda} . \\ 
        {\bf ARAE}: & it 's the \textcolor{blue}{best to eat} and \textcolor{blue}{had a great} \textcolor{blue}{meal} .\\
        {\bf iVAE$_{\text{MI}}$}: & so i had a \textcolor{red}{huge side} and the price was great . \\
        \hfill  \vspace{0mm}  \\
        
        {\bf Input}: & i am just \textcolor{red}{not a fan} of this kind of \textcolor{red}{pizza} . \\ 
        {\bf ARAE}: & i am very \textcolor{blue}{pleased} and will definitely use \textcolor{blue}{this place} . \\
        {\bf iVAE$_{\text{MI}}$}: & i am just \textcolor{red}{a fan} of \textcolor{red}{the chicken}  \textcolor{red}{and egg roll} .
        
      \end{tabular}  
    \end{tcolorbox}
    
    \begin{tcolorbox} 
      \hspace{-5mm}
    \begin{tabular}{l l}    
        {\bf Input}: & i have eaten the \textcolor{red}{lunch}  \textcolor{red}{buffet} and it was \textcolor{red}{outstanding} ! \\ 
        {\bf ARAE}: & once again , i was \textcolor{blue}{told by the wait} and was \textcolor{blue}{seated} . \\
        {\bf iVAE$_{\text{MI}}$}: & we were \textcolor{red}{not impressed} with \textcolor{red}{the buffet} there last night . \\
         \hfill  \vspace{0mm}  \\
        
        {\bf Input}: & my favorite food is \textcolor{red}{kung pao beef} , it is \textcolor{red}{delicious} . \\ 
        {\bf ARAE}: & my \textcolor{blue}{husband} was on the \textcolor{blue}{phone} , which i tried it . \\
        {\bf iVAE$_{\text{MI}}$}: & \textcolor{red}{my chicken} was n't warm , though it is \textcolor{red}{n't delicious} . \\
        \hfill  \vspace{0mm}  \\
        
        {\bf Input}: & overall , it was a very \textcolor{red}{positive} \textcolor{red}{dining experience} . \\ 
         %\vspace{1mm}
        {\bf ARAE}: & overall , it was very \textcolor{blue}{rude and}  \textcolor{blue}{unprofessional} . \\
        {\bf iVAE$_{\text{MI}}$}: & overall , it was a nightmare of \textcolor{red}{terrible experience} .
       
	\end{tabular}
	\end{tcolorbox}
	
	\caption{Sentiment transfer on $\mathtt{Yelp}$. (Up: From negative to positive, Down: From positive to negative.)}
	\label{table:sentiment_sample}
\end{table}

\paragraph{Datasets.} 
Following~\cite{shen2017style}, the $\mathtt{Yelp}$ restaurant reviews dataset is processed from the original $\mathtt{Yelp}$ dataset in language modeling. Reviews with user rating above three are considered {\em positive}, and those below three are considered {\em negative}. The pre-processing allows sentiment analysis on sentence level with feasible sentiment, ending up with shorter sentences with each at most 15 words than those in language modeling. Finally, we get two sets of unaligned reviews: 250K negative sentences, and 350K positive ones. Other dataset details are shown in Appendix \ref{data_st}.

\begin{table}[t!]
\begin{centering}
\begin{adjustbox}{scale=0.8,tabular=c|c|c|c|c|c|c,center}
\toprule
Model & Acc\! $\uparrow$ & BLEU\! $\uparrow$ & PPL\! $\downarrow$ & RPPL\! $\downarrow$ & Flu\! $\uparrow$ & Sim\! $\uparrow$ \tabularnewline
\midrule
ARAE & \textbf{95} & 32.5 & 6.8 & 395 & 3.6 & 3.5\tabularnewline
iVAE$_{\text{MI}}$ & 92 & \textbf{36.7} & \textbf{6.2} & \textbf{285} & \textbf{3.8} & \textbf{3.9}\tabularnewline
\bottomrule
\end{adjustbox}
\end{centering}
\caption{Sentiment Transfer on $\mathtt{Yelp}$.}
\label{table:sentiment}
\vspace{-3mm}
\end{table}

\paragraph{Evaluation Metrics.} (1) Acc: the accuracy of transferring sentences into another sentiment measured by an automatic classifier: the ``fasttext'' library \cite{joulin2017bag}; (2) BLEU: the consistency between the transferred text and the original; (3) PPL: the reconstruction perplexity of original sentences without altering sentiment; (4) RPPL: the reverse perplexity that evaluates the training corpus based on language model derived from the generated text, which measures the extent to which the generations are representative of the training corpus; (5) Flu: human evaluated index on the fluency of transferred sentences when read alone (1-5, 5 being
most fluent as natural language); (6) Sim: the human evaluated similarity between the original and the transferred sentences in terms of their contents (1-5, 5 being most similar). Note that the similarity measure doesn't care sentiment but only the topic covered by sentences. For human evaluation, we show 1000 randomly selected pairs of  original and transferred sentences to crowdsourcing readers, and ask them to evaluate the "Flu" and "Sim" metrics stated above. Each measure is averaged among crowdsourcing readers. 

As shown in Table \ref{table:sentiment}, iVAE$_{\text{MI}}$ outperforms ARAE
in metrics except Acc, showing that iVAE$_{\text{MI}}$ captures informative representations, generates consistently opposite sentences with similar grammatical structure and reserved semantic meaning. Both methods perform successful sentiment transfer as shown by Acc values. iVAE$_{\text{MI}}$ achieves a little lower Acc due to much more content reserving, even word reserving, of the source sentences. 

Table \ref{table:sentiment_sample} presents some examples. In each box, we show the source sentence, the transferred sentence by ARAE and iVAE$_{\text{MI}}$, respectively. We observe that ARAE usually generates new sentences that miss the content of the source, while iVAE$_{\text{MI}}$ shows better content-preserving.

\subsection{Dialog response generation}
We consider the open-domain dialog response generation task, where we need to generate a natural language response given a dialog history.
%ich represents a scenario to generate natural language based on some conversational context.
%
It is crucial to learn a meaningful latent feature representation of the dialog history in order to generate a consistent, relevant, and contentful response that is likely to drive the conversation \cite{gaosurvey}.
%on this task, as it represents semantics of next response, as well as other relevant information such as sentiment and tone.

% Conditional variational autoencoders (CVAE) \cite{zhao2017learning} serves as an important baseline, which generates responses conditioned on both context and a latent code. 

\paragraph{Datasets.}
We consider two mainstream datasets in recent studies \cite{zhao2017learning, zhao2018unsupervised, fu2019cyclical, gu2018dialogwae}: $\mathtt{Switchboard}$ \cite{godfrey1997switchboard} and $\mathtt{Dailydialog}$ \cite{li2017dailydialog}. $\mathtt{Switchboard}$ contains 2,400 two-way telephone conversations under 70 specified topics. $\mathtt{Dailydialog}$ has 13,118 daily conversations for a English learner. We process each utterance as the response of previous 10 context utterances from both speakers. The datasets are separated into training, validation, and test sets as convention: 2316:60:62 for $\mathtt{Switchboard}$ and 10:1:1 for $\mathtt{Dailydialog}$, respectively.

\paragraph{Model Extension.}
We adapt iVAE$_{\text{MI}}$ by integrating the context embedding $\boldsymbol{c}$ into all model components. The prior $p(\zv|\boldsymbol{c})$ is defined as an implicit mapping between context embedding $\boldsymbol{c}$ and prior samples, which is not pre-fixed but learned together with the variational posterior for more modeling flexibility. The encoder $q(\zv|\xv,\boldsymbol{c})$, auxiliary dual function $\nu_{\psiv}(\zv,\boldsymbol{c})$ and decoder $p(\xv|\boldsymbol{c},\zv)$ depend on context embedding $\boldsymbol{c}$ as well. Both encoder and decoder are implemented as GRUs. The utterance encoder is a bidirectional GRU with 300 hidden units in each direction. The context encoder and decoder are both GRUs with 300 hidden units. Appendix \ref{train_dial} presents more training details. 

\paragraph{Baseline.}
We benchmark representative baselines and state-of-the-art approaches, include: SeqGAN, a GAN based model for sequence generation \cite{li2017adversarial}; CVAE baseline \cite{zhao2017learning}; dialogue WAE, a conditional Wasserstein auto-encoder for response generation \cite{gu2018dialogwae}.

\begin{table}[t]
\begin{centering}
\begin{adjustbox}{scale=0.9,tabular=c|c|c|c|c,center}
\toprule
Metrics & SeqGAN & CVAE & WAE & iVAE$_{\text{MI}}$\tabularnewline
\hline
\multicolumn{5}{c}{Dataset: $\mathtt{Switchboard}$}\tabularnewline
\hline
BLEU-R\! $\uparrow$ & 0.282 & 0.295 & 0.394 & \textbf{0.427}\tabularnewline
BLEU-P\! $\uparrow$ & \textbf{0.282} & 0.258 & 0.254 & 0.254\tabularnewline
BLEU-F1\! $\uparrow$ & 0.282 & 0.275 & 0.309 & \textbf{0.319}\tabularnewline
BOW-A\! $\uparrow$ & 0.817 & 0.836 & 0.897 & \textbf{0.930}\tabularnewline
BOW-E\! $\uparrow$ & 0.515 & 0.572 & 0.627 & \textbf{0.670}\tabularnewline
BOW-G\! $\uparrow$ & 0.748 & 0.846 & 0.887 & \textbf{0.900}\tabularnewline
Intra-dist1\! $\uparrow$ & 0.705 & 0.803 & 0.713 & \textbf{0.828}\tabularnewline
Intra-dist2\! $\uparrow$ & 0.521 & 0.415 & 0.651 & \textbf{0.692}\tabularnewline
Inter-dist1\! $\uparrow$ & 0.070 & 0.112 & 0.245 & \textbf{0.391}\tabularnewline
Inter-dist2\! $\uparrow$ & 0.052 & 0.102 & 0.413 & \textbf{0.668}\tabularnewline
\hline
\multicolumn{5}{c}{Dataset: $\mathtt{Dailydialog}$}\tabularnewline
\hline
BLEU-R\! $\uparrow$ & 0.270 & 0.265 & 0.341 & \textbf{0.355}\tabularnewline
BLEU-P\! $\uparrow$ & 0.270 & 0.222 & \textbf{0.278} & 0.239\tabularnewline
BLEU-F1\! $\uparrow$ & 0.270 & 0.242 & \textbf{0.306} & 0.285\tabularnewline
BOW-A\! $\uparrow$ & 0.907 & 0.923 & 0.948 & \textbf{0.951}\tabularnewline
BOW-E\! $\uparrow$ & 0.495 & 0.543 & 0.578 & \textbf{0.609}\tabularnewline
BOW-G\! $\uparrow$ & 0.774 & 0.811 & 0.846 & \textbf{0.872}\tabularnewline
Intra-dist1\! $\uparrow$ & 0.747 & \textbf{0.938} & 0.830 & 0.897\tabularnewline
Intra-dist2\! $\uparrow$ & 0.806 & 0.973 & 0.940 & \textbf{0.975}\tabularnewline
Inter-dist1\! $\uparrow$ & 0.075 & 0.177 & 0.327 & \textbf{0.501}\tabularnewline
Inter-dist2\! $\uparrow$ & 0.081 & 0.222 & 0.583 & \textbf{0.868}\tabularnewline
\bottomrule
\end{adjustbox}
\end{centering}
\caption{Dialog response generation on two datasets. }
\label{table:dialogue}
\vspace{-0.3cm}
\end{table}

\paragraph{Evaluation Metrics.}
We adopt several widely used numerical metrics to systematically evaluate the response generation, including BLEU score \cite{papineni2002bleu}, BOW Embedding \cite{liu2016not} and Distinct \cite{li2015diversity}, as used in \cite{gu2018dialogwae}. For each testing context, we sample 10 responses from each model.
\begin{itemize}
    \item BLEU measures how much a generated response contains n-gram overlaps with the references. We compute 4-gram BLEU. For each test context, we sample 10 responses from the models and compute their BLEU scores. BLEU precision and recall is defind as the average and maximum scores, respectively \cite{zhao2017learning}.
    \vspace{-2mm}
    \item BOW embedding is the cosine similarity of bag-of-words embeddings between the generations and the reference. We compute three different BOW embedding: greedy, average, and extreme. 
    \item Distinct evaluates the diversity of the generated responses: dist-n is defined as the ratio of unique n-grams (n=1,2) over all n-grams in the generated responses. We evaluate diversities for both within each sampled response and among all responses as intra-dist and inter-dist, respectively. 
\end{itemize}

Tables \ref{table:dialogue} show the performance comparison. iVAE$_{\text{MI}}$ achieves consistent improvement on a majority of the metrics. Especially, the BOW embeddings and Distinct get significantly improvement, which implies that iVAE$_{\text{MI}}$ produces both meaningful and diverse latent representations. 

% In practice, our model is easier to tune, faster to converge compared with other adversarially trained method, such as WAE.

% \subsection{Summary of experiments} 

\section{Conclusion}
We present two types of implicit deep latent variable models, iVAE and iVAE$_{\text{MI}}$.
Core to these two models is the sample-based representation of the latent features in LVM, in replacement of traditional Gaussian-based distributions. Extensive experiments show that the proposed implicit LVM models consistently outperform the vanilla VAEs on three tasks, including language modeling, style transfer and dialog response generation. 

% Our experiments consistently show that our implicit LVMs better manipulate the latent space and produce more flexible, meaningful and diverse representations than other VAE variants. It also significantly alleviates the posterior collapse issue.

% Future work includes improving the interpretability of code samples, for instance, introducing discrete codes into the framework.

% \newpage

\subsection*{Acknowledgments}
We acknowledge Yufan Zhou for helpful discussions and the reviewers for their comments. The authors appreciate some peer researchers for discussing with us the approximation nature of our metrics, such as perplexity and KL terms, among whom are Leo Long and Liqun Chen.

\bibliography{emnlp2019}
\bibliographystyle{acl_natbib}

\appendix
\clearpage{} 
\begin{center}\LARGE{Appendix}\end{center}

\section{Decomposition of the regularization term}\label{decomp}
We have the following decomposition
\[
\begin{aligned} & -\KL\left(q_{\phiv}(\zv)\parallel p(\zv)\right)\\
 & =-\mathbb{E}_{\zv\sim q_{\phiv}(\zv)}\left[\text{log}q_{\phiv}(\zv)-\text{log}p(\zv)\right]\\
 & =H\left(q_{\phiv}(\zv)\right)+\mathbb{E}_{\zv\sim q_{\phiv}(\zv)}\text{log}p(\zv)\\
 & =H\left(q_{\phiv}(\zv)\right)+\mathbb{E}_{\xv}\mathbb{E}_{\zv\sim q_{\phiv}(\zv|\xv)}\text{log}q_{\phiv}(\zv|\xv)\\
 & -\mathbb{E}_{\xv}\mathbb{E}_{\zv\sim q_{\phiv}(\zv|\xv)}\text{log}q_{\phiv}(\zv|\xv)+\mathbb{E}_{\zv\sim q_{\phiv}(\zv)}\text{log}p(\zv)\\
 & =H\left(q_{\phiv}(\zv)\right)-H\left(q_{\phiv}(\zv|\xv)\right)\\
 & -\mathbb{E}_{\xv}\mathbb{E}_{\zv\sim q_{\phiv}(\zv|\xv)}\text{log}q_{\phiv}(\zv|\xv)+\mathbb{E}_{\zv\sim q_{\phiv}(\zv)}\text{log}p(\zv)\\
 & =I(\xv,\zv)-\mathbb{E}_{\xv}\mathbb{E}_{\zv\sim q_{\phiv}(\zv|\xv)}\text{log}q_{\phiv}(\zv|\xv)\\
 & +\mathbb{E}_{\xv}\mathbb{E}_{\zv\sim q_{\phiv}(\zv|\xv)}\text{log}p(\zv)\\
 & =I(\xv,\zv)-\mathbb{E}_{\xv}\KL\left(q_{\phiv}(\zv|\xv)\parallel p(\zv)\right)
\end{aligned}
\]
where $q_{\phiv}(\zv) = \int q(\xv)q_{\phiv}(\zv|\xv)dx$$= \mathbb{E}_{\xv}\left[q_{\phiv}(\zv|\xv)\right]$, $q(\xv)$ is the empirical data distribution for the training dataset $\mathcal{D}$; $H(\cdot)$ is the Shannon entropy and $I(\xv,\zv)=H(\zv)-H(\zv|\xv)$ is the mutual information between $z$ and $x$ under the joint distribution $q_{\phiv}(\xv,\zv)=q(\xv)q_{\phiv}(\zv|\xv)$.

\section{Training scheme of iVAE$_{\text{MI}}$}\label{I-VAE-MI}
The full training scheme of iVAE$_{\text{MI}}$ is as following after substituting the auxiliary network of implicit VAE:
\begin{itemize}
\item Sample mini batch $\xv_i\sim\mathcal{D}$, $\epsilon_i\sim q(\epsilon)$, $\zv_i\sim p(\zv)$ and generate $\zv_{\xv_i,\epsilon_i} = G(\xv_i, \epsilon_i; \phiv)$;
\item Update $\nu_{\psiv}(\zv)$ net to maximize $\sum_{i}\nu_{\psiv}(\zv_{\xv_i,\epsilon_i})$-$\sum_{i}\text{exp}(\nu_{\psiv}(\zv_i))$ according to Eq. \eqref{eq:I-VAE-MI};
\item Update parameters $\phiv$, $\thetav$ to maximize $\sum_{i}\text{log}p_{\thetav}(\xv_i|\zv_{\xv_i,\epsilon_i})$-$\sum_{i}\nu_{\psiv}(\zv_{\xv_i,\epsilon_i})$ according to Eq. \eqref{eq:I-VAE-MI}.
\end{itemize}

\section{Experimental details}

\subsection{Language modeling}\label{exp_lm}

\subsubsection{Data details}\label{data_lm}
The details of the three datasets are shown below. Each datasets contain ``train'', ``validate'' and ``test'' text files with each sentence in one line.

\subsubsection{More training details}\label{train_lm}
We basically follow the architecture specifications in \cite{kim2018semi}. They have an encoder predicting posterior mean and variance and we have an implicit encoder which predicts code samples directly from LSTM encoding and Gaussian noise. 

We implement both encoder and decoder as one-layer LSTMs with 1024 hidden units and 512-dimensional word embeddings (in $\mathtt{PTB}$, these numbers are 512 and 256). The $\nu$ network is a fully connected network whose hidden units are 512-512-512-1. The vocabulary size is 20K for $\mathtt{Yahoo}$ and $\mathtt{Yelp}$ and 10K for $\mathtt{PTB}$. The last hidden state of the encoder concatenated with a 32-dimensional Gaussian noise is used to sample 32-dimensional latent codes, which is then adopted to predict the initial hidden state of the decoder LSTM and additionally fed as input to each step at the LSTM decoder. A KL-cost annealing strategy is commonly used \cite{bowman2015generating}, where the scalar weight on the KL term is increased linearly from 0.1 to 1.0 each batch over 10 epochs. There are dropout layers with probability 0.5 between the input-to-hidden layer and the hidden-to-output layer on the decoder LSTM only.

All models are trained with Adam optimizer, when encoder and decoder LSTM use learning rate 1e-3 or 8e-4, and $\nu$ network uses learning rate 1e-5 and usually update 5 times versus a single LSTM end-to-end update. We train for 30$\sim$40 epochs, which is enough for convergence for all models. Model parameters are initialized over $U(-0.01, 0.01)$ and gradients are clipped at 5.0.

SA-VAE runs with 10 refinement steps.

\begin{comment}
\subsection{Unsupervised language pre-training}\label{exp_ulpt}
\subsubsection{Data details}\label{data_ulpt}
We use a $\mathtt{Yelp}$ restaurant reviews dataset processed from the original $\mathtt{Yelp}$ dataset in language modeling. Reviews with user rating above three are considered positive, and those below three are considered negative. Hence, this is a binary classification problem. The pre-processing in \cite{shen2017style} allows sentiment analysis on sentence level and sentences are relatively shorter than that in language modeling. It filters the sentences by eliminating those that exceed 15 words. The resulting dataset has 250K negative sentences, and 350K positive ones. The vocabulary size is 10K after replacing words occurring less than 5 times with the ``$<$unk$>$'' token. They are then divided in a ratio 7:1:2 as train, validate, and test data.
\subsubsection{More training details}\label{train_ulpt}
We use same one-layer LSTM decoders as in language modeling with 256 hidden units and 256-dimensional word embeddings. The $\nu$ networks are same as well. The encoders are all implemented with convolution layers. Each sentence as a list of word embeddings is passed through one-by-one three 2D convolution layers with respectively [2,4,32] output channels, [3,3,3] height kernel size and [2,2,2] stride sizes. Batch normalization is performed between convolution layers. The  final convoluted output is a 32 dimensional feature vector, which is passed to a fully connected layer for finally 32 dimensional latent code.
\end{comment}

\subsection{Unsupervised style transfer}\label{exp_st}
\subsubsection{Processing details}\label{data_st}
The pre-processing allows sentiment analysis on sentence level with feasible sentiment, ending up with shorter sentences with each at most 15 words than those in language modeling. Finally, we get two sets of unaligned reviews: 250K negative sentences, and 350K positive ones. They are then divided in ratio 7:1:2 to train, validate, and test data.

\subsubsection{More training details}\label{train_st}
The vocabulary size is 10K after replacing words occurring less than 5 times with the ``$<$unk$>$'' token. The encoder and decoder are all an one-layer LSTM with 128 hidden units. The word embedding size is 128 with latent content code dimension 32. When LSTMs are trained with SGD optimizer with 1.0 learning rate, other fully connected neural nets are all trained with Adam optimizer with initial learning rate 1e-4. A grad clipping on the LSTMs is performed before each update, with max grad norm set to 1. The prior net, $\nu$ auxiliary net, and style classifier are all implemented with 2 hidden layer (not include output layer) with 128-128 hidden units. The $\nu$ network is trained 5 iterations versus other networks with single iteration with each mini-batch for performing accurate regularization. The ``fasttext'' python library \cite{joulin2017bag} is needed to evaluate accuracy, and ``KenLM Language Model Toolkit'' \cite{Heafield-estimate} is needed to evaluate reverse PPL.

\subsection{Dialog response generation}\label{exp_dial}

\subsubsection{More training details}\label{train_dial}
Both encoder and decoder are implemented as GRUs. The utterance encoder is a bidirectional GRU with 300 hidden units in each direction. The context encoder and decoder are both GRUs with 300 hidden units. The prior and the recognition networks (implicit mappings) are both 2-layer feed-forward networks of size 200 with tanh non-linearity to sample Gaussian noise plus 3-layer feed-forward networks to implicitly generate samples with ReLU non-linearity and hidden sizes of 200, 200 and 400, respectively. The dimension of a latent variable z is set to 200. The initial weights for all fully connected layers are sampled from a uniform distribution [-0.02, 0.02]. We set the vocabulary size to 10,000 and define all the out-of-vocabulary words to a special token ``$<$unk$>$''. The word embedding size is 200 and initialized with pre-trained Glove vectors \cite{pennington2014glove} on Twitter. The size of context window is set to 10 with a maximum utterance length of 40. We sample responses with greedy decoding so that the randomness entirely come from the latent variables. The GRUs are trained by SGD optimizer with an initial learning rate of 1.0, decay rate 0.4 and gradient clipping at 1. The fully connected neural nets are trained by RMSprop optimizer with fixed learning rates of 5e-5, except the $\nu$ net is trained with learning rate 1e-5 and repeatedly 5 inner iterations.

\end{document}

%% file: subtex/mlVecMat.tex
\usepackage{verbatim}

\usepackage{anyfontsize}

\usepackage{color}
\usepackage{tikz}
\usetikzlibrary{arrows,shapes,decorations,automata,backgrounds,fit,petri}
\usepackage{adjustbox}

\newcommand{\RN}[1]{%
	\textup{\lowercase\expandafter{\it \romannumeral#1}}%
}

\makeatletter
\newcommand{\distas}[1]{\mathbin{\overset{#1}{\kern\z@\sim}}}%

\usepackage{enumitem}

\newcommand{\ie}[0]{\emph{i.e., }}

\newcommand{\beq}{\vspace{0mm}\begin{equation}}
\newcommand{\eeq}{\vspace{0mm}\end{equation}}
\newcommand{\beqs}{\vspace{0mm}\begin{eqnarray}}
\newcommand{\eeqs}{\vspace{0mm}\end{eqnarray}}
\newcommand{\barr}{\begin{array}}
\newcommand{\earr}{\end{array}}

\newcommand{\hv}[0]{{\boldsymbol{h}}}

\newcommand{\xv}{\boldsymbol{x}}

\newcommand{\zv}{\boldsymbol{z}}

\newcommand{\thetav}{\boldsymbol{\theta}}

\newcommand{\phiv}{\boldsymbol{\phi}}
\newcommand{\psiv}{\boldsymbol{\psi}}

\newcommand{\Lcal}{\mathcal{L}}
\newcommand{\Ncal}{\mathcal{N}}

\newcommand{\Dcal}{\mathcal{D}}

%% file: emnlp2019.bbl
\begin{thebibliography}{43}
\expandafter\ifx\csname natexlab\endcsname\relax\def\natexlab#1{#1}\fi

\bibitem[{Arjovsky et~al.(2017)Arjovsky, Chintala, and
  Bottou}]{Arjovsky2017WassersteinG}
Mart{\'\i}n Arjovsky, Soumith Chintala, and L{\'e}on Bottou. 2017.
\newblock Wasserstein gan.
\newblock \emph{CoRR}, abs/1701.07875.

\bibitem[{Bowman et~al.(2015)Bowman, Vilnis, Vinyals, Dai, Jozefowicz, and
  Bengio}]{bowman2015generating}
Samuel~R Bowman, Luke Vilnis, Oriol Vinyals, Andrew~M Dai, Rafal Jozefowicz,
  and Samy Bengio. 2015.
\newblock Generating sentences from a continuous space.
\newblock \emph{arXiv preprint arXiv:1511.06349}.

\bibitem[{Chen et~al.(2018)Chen, Li, Chen, Wang, Pu, and
  Carin}]{chen2018continuous}
Changyou Chen, Chunyuan Li, Liqun Chen, Wenlin Wang, Yunchen Pu, and Lawrence
  Carin. 2018.
\newblock Continuous-time flows for efficient inference and density estimation.
\newblock \emph{ICML}.

\bibitem[{Cremer et~al.(2018)Cremer, Li, and Duvenaud}]{cremer2018inference}
Chris Cremer, Xuechen Li, and David Duvenaud. 2018.
\newblock Inference suboptimality in variational autoencoders.
\newblock \emph{arXiv preprint arXiv:1801.03558}.

\bibitem[{Dai et~al.(2018)Dai, Dai, He, Liu, Liu, Chen, Xiao, and
  Song}]{NIPS2018_8177}
Bo~Dai, Hanjun Dai, Niao He, Weiyang Liu, Zhen Liu, Jianshu Chen, Lin Xiao, and
  Le~Song. 2018.
\newblock Coupled variational bayes via optimization embedding.
\newblock In S.~Bengio, H.~Wallach, H.~Larochelle, K.~Grauman, N.~Cesa-Bianchi,
  and R.~Garnett, editors, \emph{Advances in Neural Information Processing
  Systems 31}, pages 9713--9723. Curran Associates, Inc.

\bibitem[{Donahue et~al.(2017)Donahue, Kr{\"a}henb{\"u}hl, and
  Darrell}]{donahue2017adversarial}
Jeff Donahue, Philipp Kr{\"a}henb{\"u}hl, and Trevor Darrell. 2017.
\newblock Adversarial feature learning.
\newblock \emph{ICLR}.

\bibitem[{Fu et~al.(2019)Fu, Li, Liu, Gao, Celikyilmaz, and
  Carin}]{fu2019cyclical}
Hao Fu, Chunyuan Li, Xiaodong Liu, Jianfeng Gao, Asli Celikyilmaz, and Lawrence
  Carin. 2019.
\newblock Cyclical annealing schedule: A simple approach to mitigating kl
  vanishing.
\newblock \emph{NAACL}.

\bibitem[{Gao et~al.(2019)Gao, Galley, and Li}]{gaosurvey}
Jianfeng Gao, Michel Galley, and Lihong Li. 2019.
\newblock Neural approaches to conversational ai.
\newblock \emph{Foundations and Trends{\textregistered} in Information
  Retrieval}, 13(2-3):127--298.

\bibitem[{Godfrey and Holliman(1997)}]{godfrey1997switchboard}
John~J Godfrey and Edward Holliman. 1997.
\newblock Switchboard-1 release 2.
\newblock \emph{Linguistic Data Consortium, Philadelphia}, 926:927.

\bibitem[{Goodfellow et~al.(2014)Goodfellow, Pouget-Abadie, Mirza, Xu,
  Warde-Farley, Ozair, Courville, and Bengio}]{GAN}
Ian Goodfellow, Jean Pouget-Abadie, Mehdi Mirza, Bing Xu, David Warde-Farley,
  Sherjil Ozair, Aaron Courville, and Yoshua Bengio. 2014.
\newblock \href
  {http://papers.nips.cc/paper/5423-generative-adversarial-nets.pdf}
  {Generative adversarial nets}.
\newblock In Z.~Ghahramani, M.~Welling, C.~Cortes, N.~D. Lawrence, and K.~Q.
  Weinberger, editors, \emph{Advances in Neural Information Processing Systems
  27}, pages 2672--2680. Curran Associates, Inc.

\bibitem[{Gu et~al.(2018)Gu, Cho, Ha, and Kim}]{gu2018dialogwae}
Xiaodong Gu, Kyunghyun Cho, Jungwoo Ha, and Sunghun Kim. 2018.
\newblock Dialogwae: Multimodal response generation with conditional
  wasserstein auto-encoder.
\newblock \emph{arXiv preprint arXiv:1805.12352}.

\bibitem[{Gulrajani et~al.(2017)Gulrajani, Ahmed, Arjovsky, Dumoulin, and
  Courville}]{gulrajani2017wasserstein}
Ishaan Gulrajani, Faruk Ahmed, Martin Arjovsky, Vincent Dumoulin, and Aaron
  Courville. 2017.
\newblock \href
  {https://papers.nips.cc/paper/7159-improved-training-of-wasserstein-gans}
  {Improved training of wasserstein gans}.
\newblock In \emph{Advances in Neural Information Processing Systems 30}, pages
  5769--5779.
\newblock Arxiv: 1704.00028.

\bibitem[{He et~al.(2019)He, Spokoyny, Neubig, and
  Berg-Kirkpatrick}]{he2019lagging}
Junxian He, Daniel Spokoyny, Graham Neubig, and Taylor Berg-Kirkpatrick. 2019.
\newblock Lagging inference networks and posterior collapse in variational
  autoencoders.
\newblock \emph{arXiv preprint arXiv:1901.05534}.

\bibitem[{Heafield et~al.(2013)Heafield, Pouzyrevsky, Clark, and
  Koehn}]{Heafield-estimate}
Kenneth Heafield, Ivan Pouzyrevsky, Jonathan~H. Clark, and Philipp Koehn. 2013.
\newblock \href {https://kheafield.com/papers/edinburgh/estimate\_paper.pdf}
  {Scalable modified {Kneser-Ney} language model estimation}.
\newblock In \emph{Proceedings of the 51st Annual Meeting of the Association
  for Computational Linguistics}, pages 690--696, Sofia, Bulgaria.

\bibitem[{Higgins et~al.(2017)Higgins, Matthey, Pal, Burgess, Glorot,
  Botvinick, Mohamed, and Lerchner}]{higgins2017beta}
Irina Higgins, Loic Matthey, Arka Pal, Christopher Burgess, Xavier Glorot,
  Matthew Botvinick, Shakir Mohamed, and Alexander Lerchner. 2017.
\newblock beta-vae: Learning basic visual concepts with a constrained
  variational framework.
\newblock In \emph{International Conference on Learning Representations},
  volume~3.

\bibitem[{Hochreiter and Schmidhuber(1997)}]{hochreiter1997long}
Sepp Hochreiter and J{\"u}rgen Schmidhuber. 1997.
\newblock Long short-term memory.
\newblock \emph{Neural computation}, 9(8):1735--1780.

\bibitem[{Hu et~al.(2017)Hu, Yang, Liang, Salakhutdinov, and
  Xing}]{hu2017toward}
Zhiting Hu, Zichao Yang, Xiaodan Liang, Ruslan Salakhutdinov, and Eric~P Xing.
  2017.
\newblock Toward controlled generation of text.
\newblock In \emph{Proceedings of the 34th International Conference on Machine
  Learning-Volume 70}, pages 1587--1596. JMLR. org.

\bibitem[{Joulin et~al.(2017)Joulin, Grave, Bojanowski, and
  Mikolov}]{joulin2017bag}
Armand Joulin, Edouard Grave, Piotr Bojanowski, and Tomas Mikolov. 2017.
\newblock Bag of tricks for efficient text classification.
\newblock In \emph{Proceedings of the 15th Conference of the European Chapter
  of the Association for Computational Linguistics: Volume 2, Short Papers},
  pages 427--431. Association for Computational Linguistics.

\bibitem[{Kim et~al.(2018)Kim, Wiseman, Miller, Sontag, and Rush}]{kim2018semi}
Yoon Kim, Sam Wiseman, Andrew Miller, David Sontag, and Alexander Rush. 2018.
\newblock Semi-amortized variational autoencoders.
\newblock In \emph{International Conference on Machine Learning}, pages
  2683--2692.

\bibitem[{Kim et~al.(2017)Kim, Zhang, Rush, LeCun
  et~al.}]{kim2017adversarially}
Yoon Kim, Kelly Zhang, Alexander~M Rush, Yann LeCun, et~al. 2017.
\newblock Adversarially regularized autoencoders.
\newblock \emph{arXiv preprint arXiv:1706.04223}.

\bibitem[{Kingma and Welling(2013)}]{kingma2013auto}
Diederik~P Kingma and Max Welling. 2013.
\newblock Auto-encoding variational bayes.
\newblock \emph{arXiv preprint arXiv:1312.6114}.

\bibitem[{Li et~al.(2017{\natexlab{a}})Li, Liu, Chen, Pu, Chen, Henao, and
  Carin}]{li2017alice}
Chunyuan Li, Hao Liu, Changyou Chen, Yuchen Pu, Liqun Chen, Ricardo Henao, and
  Lawrence Carin. 2017{\natexlab{a}}.
\newblock Alice: Towards understanding adversarial learning for joint
  distribution matching.
\newblock In \emph{Advances in Neural Information Processing Systems}, pages
  5495--5503.

\bibitem[{Li et~al.(2015)Li, Galley, Brockett, Gao, and
  Dolan}]{li2015diversity}
Jiwei Li, Michel Galley, Chris Brockett, Jianfeng Gao, and Bill Dolan. 2015.
\newblock A diversity-promoting objective function for neural conversation
  models.
\newblock \emph{arXiv preprint arXiv:1510.03055}.

\bibitem[{Li et~al.(2017{\natexlab{b}})Li, Monroe, Shi, Jean, Ritter, and
  Jurafsky}]{li2017adversarial}
Jiwei Li, Will Monroe, Tianlin Shi, S{\'e}bastien Jean, Alan Ritter, and Dan
  Jurafsky. 2017{\natexlab{b}}.
\newblock Adversarial learning for neural dialogue generation.
\newblock \emph{arXiv preprint arXiv:1701.06547}.

\bibitem[{Li et~al.(2017{\natexlab{c}})Li, Su, Shen, Li, Cao, and
  Niu}]{li2017dailydialog}
Yanran Li, Hui Su, Xiaoyu Shen, Wenjie Li, Ziqiang Cao, and Shuzi Niu.
  2017{\natexlab{c}}.
\newblock Dailydialog: A manually labelled multi-turn dialogue dataset.
\newblock \emph{arXiv preprint arXiv:1710.03957}.

\bibitem[{Li et~al.(2017{\natexlab{d}})Li, Turner, and Liu}]{li2017approximate}
Yingzhen Li, Richard~E Turner, and Qiang Liu. 2017{\natexlab{d}}.
\newblock Approximate inference with amortised mcmc.
\newblock \emph{arXiv preprint arXiv:1702.08343}.

\bibitem[{Liu et~al.(2016)Liu, Lowe, Serban, Noseworthy, Charlin, and
  Pineau}]{liu2016not}
Chia-Wei Liu, Ryan Lowe, Iulian~V Serban, Michael Noseworthy, Laurent Charlin,
  and Joelle Pineau. 2016.
\newblock How not to evaluate your dialogue system: An empirical study of
  unsupervised evaluation metrics for dialogue response generation.
\newblock \emph{arXiv preprint arXiv:1603.08023}.

\bibitem[{Makhzani et~al.(2015)Makhzani, Shlens, Jaitly, Goodfellow, and
  Frey}]{makhzani2015adversarial}
Alireza Makhzani, Jonathon Shlens, Navdeep Jaitly, Ian Goodfellow, and Brendan
  Frey. 2015.
\newblock Adversarial autoencoders.
\newblock \emph{arXiv preprint arXiv:1511.05644}.

\bibitem[{Marcus et~al.(1993)Marcus, Santorini, and
  Marcinkiewicz}]{marcus1993building}
Mitchell Marcus, Beatrice Santorini, and Mary~Ann Marcinkiewicz. 1993.
\newblock Building a large annotated corpus of english: The penn treebank.
\newblock \emph{Computational Linguistics}.

\bibitem[{Mescheder et~al.(2017)Mescheder, Nowozin, and
  Geiger}]{mescheder2017adversarial}
Lars Mescheder, Sebastian Nowozin, and Andreas Geiger. 2017.
\newblock Adversarial variational bayes: Unifying variational autoencoders and
  generative adversarial networks.
\newblock In \emph{Proceedings of the 34th International Conference on Machine
  Learning-Volume 70}, pages 2391--2400. JMLR. org.

\bibitem[{Miao et~al.(2016)Miao, Yu, and Blunsom}]{miao2016neural}
Yishu Miao, Lei Yu, and Phil Blunsom. 2016.
\newblock Neural variational inference for text processing.
\newblock In \emph{International conference on machine learning}, pages
  1727--1736.

\bibitem[{Papineni et~al.(2002)Papineni, Roukos, Ward, and
  Zhu}]{papineni2002bleu}
Kishore Papineni, Salim Roukos, Todd Ward, and Wei-Jing Zhu. 2002.
\newblock Bleu: a method for automatic evaluation of machine translation.
\newblock In \emph{Proceedings of the 40th annual meeting on association for
  computational linguistics}, pages 311--318. Association for Computational
  Linguistics.

\bibitem[{Pennington et~al.(2014)Pennington, Socher, and
  Manning}]{pennington2014glove}
Jeffrey Pennington, Richard Socher, and Christopher Manning. 2014.
\newblock Glove: Global vectors for word representation.
\newblock In \emph{Proceedings of the 2014 conference on empirical methods in
  natural language processing (EMNLP)}, pages 1532--1543.

\bibitem[{Pu et~al.(2017{\natexlab{a}})Pu, Gan, Henao, Li, Han, and
  Carin}]{pu2017vae}
Yuchen Pu, Zhe Gan, Ricardo Henao, Chunyuan Li, Shaobo Han, and Lawrence Carin.
  2017{\natexlab{a}}.
\newblock Vae learning via stein variational gradient descent.
\newblock In \emph{Advances in Neural Information Processing Systems}, pages
  4236--4245.

\bibitem[{Pu et~al.(2017{\natexlab{b}})Pu, Wang, Henao, Chen, Gan, Li, and
  Carin}]{pu2017adversarial}
Yuchen Pu, Weiyao Wang, Ricardo Henao, Liqun Chen, Zhe Gan, Chunyuan Li, and
  Lawrence Carin. 2017{\natexlab{b}}.
\newblock Adversarial symmetric variational autoencoder.
\newblock In \emph{Advances in Neural Information Processing Systems}, pages
  4330--4339.

\bibitem[{Rezende et~al.(2014)Rezende, Mohamed, and
  Wierstra}]{rezende2014stochastic}
Danilo~Jimenez Rezende, Shakir Mohamed, and Daan Wierstra. 2014.
\newblock Stochastic backpropagation and approximate inference in deep
  generative models.
\newblock In \emph{International Conference on Machine Learning}, pages
  1278--1286.

\bibitem[{Rockafellar et~al.(1966)}]{rockafellar1966extension}
R~Tyrrell Rockafellar et~al. 1966.
\newblock Extension of fenchel'duality theorem for convex functions.
\newblock \emph{Duke mathematical journal}, 33(1):81--89.

\bibitem[{Shah and Barber(2018)}]{shah_generative}
Harshil Shah and David Barber. 2018.
\newblock \href
  {http://papers.nips.cc/paper/7409-generative-neural-machine-translation.pdf}
  {Generative neural machine translation}.
\newblock In S.~Bengio, H.~Wallach, H.~Larochelle, K.~Grauman, N.~Cesa-Bianchi,
  and R.~Garnett, editors, \emph{Advances in Neural Information Processing
  Systems 31}, pages 1346--1355. Curran Associates, Inc.

\bibitem[{Shen et~al.(2017)Shen, Lei, Barzilay, and Jaakkola}]{shen2017style}
Tianxiao Shen, Tao Lei, Regina Barzilay, and Tommi Jaakkola. 2017.
\newblock Style transfer from non-parallel text by cross-alignment.
\newblock In \emph{Advances in neural information processing systems}, pages
  6830--6841.

\bibitem[{Yang et~al.(2017)Yang, Hu, Salakhutdinov, and
  Berg-Kirkpatrick}]{yang2017improved}
Zichao Yang, Zhiting Hu, Ruslan Salakhutdinov, and Taylor Berg-Kirkpatrick.
  2017.
\newblock Improved variational autoencoders for text modeling using dilated
  convolutions.
\newblock In \emph{Proceedings of the 34th International Conference on Machine
  Learning-Volume 70}, pages 3881--3890. JMLR. org.

\bibitem[{Zhao et~al.(2017{\natexlab{a}})Zhao, Song, and
  Ermon}]{zhao2017infovae}
Shengjia Zhao, Jiaming Song, and Stefano Ermon. 2017{\natexlab{a}}.
\newblock Info{VAE}: Information maximizing variational autoencoders.
\newblock \emph{arXiv preprint arXiv:1706.02262}.

\bibitem[{Zhao et~al.(2018)Zhao, Lee, and Eskenazi}]{zhao2018unsupervised}
Tiancheng Zhao, Kyusong Lee, and Maxine Eskenazi. 2018.
\newblock Unsupervised discrete sentence representation learning for
  interpretable neural dialog generation.
\newblock \emph{arXiv preprint arXiv:1804.08069}.

\bibitem[{Zhao et~al.(2017{\natexlab{b}})Zhao, Zhao, and
  Eskenazi}]{zhao2017learning}
Tiancheng Zhao, Ran Zhao, and Maxine Eskenazi. 2017{\natexlab{b}}.
\newblock Learning discourse-level diversity for neural dialog models using
  conditional variational autoencoders.
\newblock \emph{arXiv preprint arXiv:1703.10960}.

\end{thebibliography}
